\pdfoutput=1

\documentclass[11pt]{article}

\usepackage[]{emnlp2021}

\usepackage{times}
\usepackage{latexsym}

\usepackage[T1]{fontenc}

\usepackage[utf8]{inputenc}

\usepackage{graphicx}
\usepackage{amsmath}
\usepackage{arydshln} 
\usepackage{CJKutf8} 

\usepackage{microtype}

\title{\emph{ Challenging Instances are Worth Learning }: \\
Generating Valuable Negative Samples for Response Selection Training}

\author{Yao Qiu, \ Jinchao Zhang, \ Huiying Ren, \ Jie Zhou \\
  Pattern Recognition Center, WeChat AI, Tencent Inc, China \\
  \texttt{\{yasinqiu, dayerzhang, feliciaren, withtomzhou\}@tencent.com}
}

\begin{document}
\maketitle

\begin{abstract}
Retrieval-based chatbot selects the appropriate response from candidates according to the context, which heavily depends on a response selection module.
A response selection module is generally a scoring model to evaluate candidates and is usually trained on the annotated positive response and sampled negative responses.
Sampling negative responses lead to two risks: 
a). The sampled negative instances, especially that from random sampling methods, are mostly irrelevant to the dialogue context and too easy to be fitted at the training stage while causing a weak model in the real scenario. 
b). The so-called negative instances may be positive, which is known as the fake negative problem.
To address the above issue, we employ pre-trained language models, such as the DialoGPT to construct more challenging negative instances to enhance the model robustness. 
Specifically, we provide garbled context to the pre-trained model to generate responses and filter the fake negative ones.
In this way, our negative instances are fluent, context-related, and more challenging for the model to learn, while can not be positive.
Extensive experiments show that our method brings significant and stable improvements on the dialogue response selection capacity.
\end{abstract}

\section{Introduction}

In recent years, building intelligent conversational agents \cite{shum2018eliza, kollar2018alexa, kim2020beyond} is gaining more and more attention in the field of natural language processing. In various types of dialogue systems, retrieval-based dialog systems \cite{lowe2015ubuntu, wu2016sequential, zhang2018modeling} are widely used in the industry because their responses are controllable, accurate, informative, and promising. In this work, we focus on multi-turn response selection task in retrieval-based dialog systems. This task aims to identify the best response from a set of candidate responses given a dialogue context, i.e., the conversation history.

\begin{table}[]
\small
\begin{tabular}{p{5cm} |  p{1.5cm}}
\hline
\textbf{Conversation History} & \textbf{Model} \\ 
 & \textbf{Prediction} \\
\hline
A: Is this made of pure cotton? &  \\
B: This dress is made of non-woven fabric. &  \\
A: Which is pure cotton? &  \\
B: The latest one in our shop is. &  \\
A: What is the difference between non-woven fabric and pure cotton? &  \\
\hline
\textbf{Golden response:} &  \\
G: Non-woven fabric and cotton are almost soft, but water absorption is slower & Positive \\
\hline
\textbf{Random sampled negative response:} &  \\
R: The order was successful, please check the shipping address. & Negative \\
\hline
\textbf{Challenging negative response:} &  \\
C1: The dress you bought is made of pure cotton. & Positive \\
C2: Non-woven clothes are out of stock now. & Positive \\
\hline
\end{tabular}
\caption{Conversation history of speakers A and B. The right column shows the predictions of the model. The model can make the correct classification on golden and random negative responses, but it makes mistakes on challenging negative responses, because these response contain history's words.}
\label{case_show}
\end{table}

For the response selection problem, most of the current practice \cite{wu2016sequential, zhou2018multi, tao2019multi, yuan2019multi} are to build utterance-response matching models based on attention mechanisms~\cite{vaswani2017attention}. These models output a score indicating the adequacy of individual response candidates in the dialogue context. Early works on this topic focus on fine-grained text encoding and better interactions between response candidates and conversation history, by specially designed matching networks \cite{wu2016sequential, zhou2016multi, lu2019constructing}.
Recently, pre-trained language models, e.g., BERT \cite{devlin2018bert}, RoBERTa \cite{liu2019roberta} and ELECTRA \cite{clark2020electra} have achieved significant performance improvements in the multi-turn response selection \cite{whang2019effective, lu2020improving, gu2020speaker, humeau2019poly}.

Previous works usually construct the negative responses by the random sampling, which has explicit limitations.
On the one hand, random sampled negative responses are usually easy to be distinguished, since they are usually irrelevant to the conversation history in terms of topic, and they can't form coherence with the dialogue context. 
On the other hand, the model trained on such naive negative responses performs poorly on challenging negative responses \cite{lin2020world}, which are similar to the conversation history. An example is given in Figure \ref{case_show}, the random sampled negative response is too easy for the model because they show little relevance to the conversation history. 
While the challenging negative responses are more confusing to the model, they mentioned some identical keywords or phrases in the dialogue context, while they have no coherence with the history. Such challenging negative responses rarely appear in the training set, but they are common in real-world scenarios.

We aim to improve the model's robustness by using challenging instances to train the model.
The challenging instances are negative responses which are more confusing than the randomly sampled negative responses. These challenging responses should be more related to the conversation history, e.g., they may have some overlap words with history, but they are not proper to be a natural response to the history.
Pre-trained language models such as DialoGPT~\cite{zhang2019dialogpt} can be used to generate responses according to the conversation history, though these responses are quite relevant to the history, most of which cannot be treated to be negative, since DialoGPT is pre-trained on large scale dialogue data, it is strong enough to generate relatively natural responses. 
To solve this problem, we design a set of conversation history garbling strategies, e.g., randomly exchange the positions of two turns of dialogue and replace some turns with random sampled utterances.
The intuition behind this is that responses generated from garbled conversation history are more likely to be negative. 
Besides, to make generated responses more confusing to the model, we insert keywords that have been mentioned in the dialogue history to the generated responses. 
To address the fake negative issue, 
we calculate the perplexity of the response given the original conversation history as the metric to select the response which is most like negative.

The proposed negative response generation method is a simple and effective data augmentation approach and it can be applied to any response selection model without any changes to the model architecture or training objective function.
Experiment results on four matching models and two benchmark datasets demonstrate that negative responses generated by our method leads to remarkable performance improvement consistently.

\section{Related Work}
\paragraph{Dialogue Responses Selection.}
In recent years, developing response selection model on multi-turn conversation has gained considerable attention \cite{lowe2015ubuntu, wu2016sequential, zhang2018modeling}.
The formulation of the response selection task is defined as follows. Given a conversation history $C$ and several candidate responses $R_i$, the goal of the matching model $M$ is to predict a score $S=M(C, R_i)$ measuring the adequacy of the candidate $R_i$ for the context $C$, then choose the candidate with the highest score as the proper response.

\paragraph{Model Architecture.}
Early works mostly focus on design various neural architectures for fine-grained context encoding and matching between conversation history and candidate response. \newcite{lowe2015ubuntu} designed a dual encoder network to better model the interaction between context and response. \newcite{wu2016sequential} leveraged several matching matrices to match context and candidate responses and proposed a model named the sequential matching network. After the effectiveness of pre-trained language models, e.g., BERT \cite{devlin2018bert} and RoBERTa\cite{liu2019roberta} based on self-attention mechanism \cite{vaswani2017attention} had been proved, subsequent works have applied it to the response selection task. \newcite{whang2019effective} first applied BERT on this task and obtained state-of-the-art performance by finetuning BERT on response selection datasets. \newcite{lu2020improving, gu2020speaker} proposed to model speaker information and showed its effectiveness.

\begin{figure*}[ht]
\centering
\includegraphics[width=14cm]{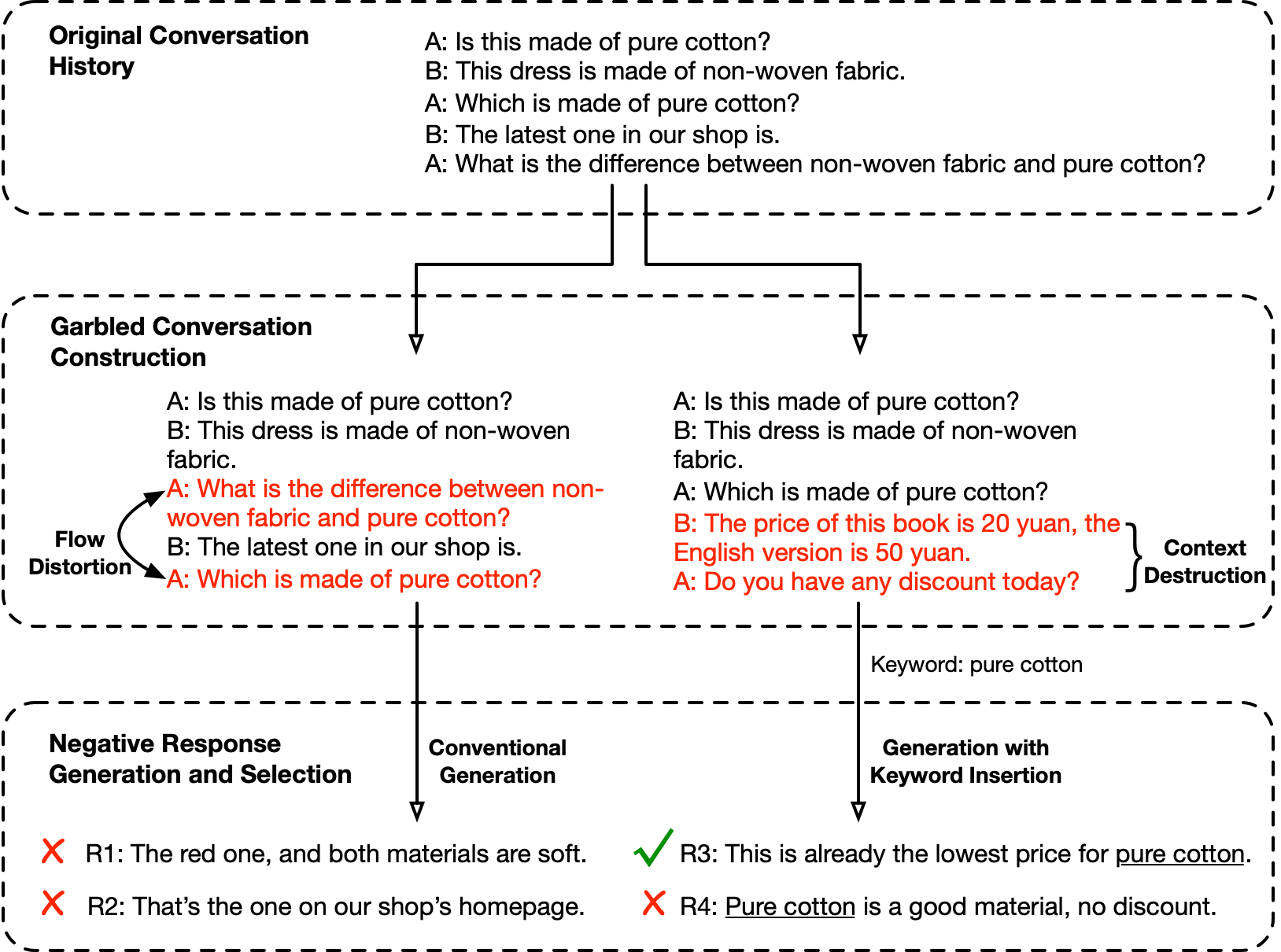}
\caption{The illustration of the challenging negative response generation. For each conversation, we first construct two garbled conversations by two strategies named flow distortion and context destruction. Then a fine-tuned DialoGPT is leveraged to generate responses based on the garbled conversations. At last, we select the response with the highest perplexity score based on the original conversation to avoid fake negatives.}
\label{main_pic}
\end{figure*}

\paragraph{Data Construction and Augmentation.}
In addition to the progress in the model architecture, \newcite{lin2020world} argue that the binary classification training objective is less effective since the quality of candidate responses can be quite diverse. 
To address this issue, they use retrieval method and generation model to automatically construct greyscale data and apply ranking objective to let the model learn this progressive relationship (\textit{ground  truth response} $>$ \textit{greyscale response} $>$ \textit{random sampled response}). However, their method is based on a strong hypothesis which is not necessarily reasonable.
Different from their work, we suppose \textit{correct is correct, wrong is wrong}, no matter how relevant response is to the conversation history, as long as it fails to form a coherent dialogue, it should be negative. 
So we still use binary classification as the training objective. 
Instead, we focus to construct more challenging negative responses for that the randomly sampled responses are too easy to be distinguished by the model. 

\section{Approach}

\subsection{Overview}
The pipeline of our method mainly consists of three steps: 
1) Fine-tuning DialoGPT on domain-specific datasets, 
2) generating challenging negative responses using DialoGPT according to the garbled conversation history, 
3) employing these generated negative instances together with an original dataset to train the response selection model. 
Step 1 and step 3 use off-the-shelf methods, so we will mainly introduce step2. Figure \ref{main_pic} depicts an overview of the negative response generation process named \textbf{GGS}, it is composed of \textbf{G}arbled Conversation Construction, Response \textbf{G}eneration and Response  \textbf{S}election. We will describe each part in detail.

\subsection{Challenging Negative Responses Definition}
We aim to generate negative responses that are more challenging than random sampled negative responses. These negative responses have one or two of the following characteristics: 
a) They must be negative, that is, they are not proper to be the response of the given conversation history according to human language habits. 
b) They should be more or less similar to the conversation history, e.g., they contain some keywords or phrases which are also mentioned in the dialogue context. Such overlap in the content may confuse the model to make incorrect predictions. 
We design the negative sample generation approach according to these characteristics.

\subsection{Garbled Conversation Construction}
State-of-the-art pre-trained language models, such as GPT2 \cite{radford2019language}, DialoGPT \cite{zhang2019dialogpt} and T5\cite{raffel2019exploring}, have been proved to be effective to generate grammatically correct and coherent response given the correct conversation history. Most of these generated responses are not proper to be treated as negative responses. 
To address this issue, we propose two conversation history garbling strategies with the intuition that a response generated from a garbled conversation history is more likely to be negative to the original history, while at the same time, it maintains the relevance with the original history.

The first strategy is called \textbf{flow distortion}, suppose the turn number of the conversation history is $N$, we randomly select an index $i \in [1, N-1]$, and exchange the position of the utterance $u_i$ and $u_N$, as shown in Figure \ref{main_pic}. By doing this, we expect the response generated by DialoGPT from this garbled history is relevant to the original conversation history in content, especially similar to the content of $turn_{i+1}$.
Because these responses are not coherent with the latest utterance, so they can be treated as negative responses. 

Another strategy is \textbf{context destruction}. In this strategy, we replace two or three latest conversation turns by randomly selected utterances in other dialogue, as shown in Figure \ref{main_pic}. By doing this, the main information of the original conversation still exists, but the topic of the conversation has suddenly changed because of the random replacing. When using DialoGPT to generate a response based on this garbled history, we expect that despite the response is about another topic, it still contains the original context's key information. Take the dialogue in Figure \ref{main_pic} as an example, though $R3$ talks about the discount, but it still mentions pure cotton which is one of the keywords in the original conversation history. To ensure the key information of the original conversation history will be mentioned in the generated response, we leverage a keyword extraction algorithm to extract keywords and insert them into the generation process, we will introduce it in detail in the next section.

\subsection{Response Generation}
Different from \newcite{lin2020world} training a response generation model from scratch, we load pre-trained language model Cdial-GPT~\cite{wang2020chinese}  and fine-tune it on task-specific datasets. We then use the fine-tuned model as our response generation model considering the responses they generate are grammatically correct and often coherent with the given conversation history. We apply two types of generation approaches.

The first generation approach is the traditional decoding method used in Cdial-GPT. This approach takes the garbled conversations constructed by flow distortion strategy as input, we first concatenate all dialogue turns into a long text $C=[x_1, x_2, ...,x_N]$, where $N$ is the sequence length. The response generation model decodes the response word by word using top-p sampling conditioned on $C$. Responses generated using this method are likely to be similar to one of the utterances in the original conversation history.

To make the generated responses more confusing, we propose the second generation approach which takes garbled conversation constructed by context destruction strategy as the input, 
we first use the traditional decoding method to generate a response $R_0 = [r_1, r_2, ..., r_K]$, at the same time, we keep a matrix $P \in [K, V]$, where $K$ is the length of generated response, and $V$ is vocabulary size, saving the probability distribution over the whole vocabulary on every decoding time step. 
Then we extract keyword or key phrase $w_{key}$ in the original conversation history, the extraction algorithm we use is CKPE~\footnote{github.com/dongrixinyu/chinese\_keyphrase\_extractor}. 
Combining $w_{key}$ and probability matrix $P$ we can find the step $i$ with the highest $w_{key}$'s probability, it is the most likely position where we can insert $w_{key}$. We replace $r_i$ with $w_{key}$ and throw $r_{i+1}, r_{i+2}, ..., r_K$ away. 
So far, we get a incomplete response $R_1 = [r_1, r_2, ..., w_{key}]$, we use Cdial-GPT to generate the rest of the response based on $R_1$ and conversation history $C$, the R2 in Figure \ref{main_pic} is an example. 
In addition to insertion, we can also simply force the response to start with $w_{key}$, the R3 in Figure \ref{main_pic} is an example.

\subsection{Response Selection}
Though we use Cdial-GPT to generate responses based on garbled conversation history, the generated responses could still be positive which is known as the fake negative problem, for the reason that responses in the open-domain dialogue are quite diverse, 
and the content of the conversation is more extensive compared to task-oriented dialogue. 
To address this issue, we propose a response selection mechanism to select the response which is more negative.

\begin{table}[]
\centering
\begin{tabular}{l lll}
\hline
 & \multicolumn{3}{c}{Douban} \\
 & Train & Val & Test \\
\hline 
Pos:Neg & 1:1 & 1:1 & 1: 9 \\
\# Dialogues & 50K & 25K & 1K \\
\# Avg turns & 8.694 & 8.751 & 8.475 \\
\hline
\hline
 & \multicolumn{3}{c}{E-Commerce} \\
 & Train & Val & Test \\
\hline 
Pos:Neg & 1:1 & 1:1 & 1:9 \\
\# Dialogues & 50K & 5K & 1K \\
\# Avg turns & 7.509 & 7.482 & 7.640 \\
\hline
\end{tabular}

\caption{Statistics of the two multi-turn response selection datasets.}
\label{dataset_info}
\end{table}

To achieve this, we directly use Cdial-GPT to calculate the perplexity of generated responses based on the original conversation history. Because Cdial-GPT is pre-trained on a large-scale dialogue dataset and fine-tuned on a task-specific dataset, it has the capacity to distinguish whether a conversation is a natural human conversation. Reflecting on the perplexity metric, the lower the value, the more likely response is a natural response. 
We denote the original conversation history as $C = x_1, x_2, ... , x_N$,
denote the generated response as $R_g = r_1, r_2, ..., r_K$, the conditional probability of $P(R_g | C)$ can be written as the product of a series of conditional probabilities:
\begin{equation}
    P(R_g|C) = \prod_{i=1}^K p(r_i | x_1,x_2,...,x_N,r_1,...,r_{i-1})
\end{equation}
The perplexity score can be calculated as:
\begin{equation}
    PPL(R_g) = P(R_g|C)^{-\frac{1}{K}}
    \label{ppl}
\end{equation}
In the response generation process, we use top-p sampling and different decoding methods to generate multiple responses $R_g^1, R_g^2,..., R_g^L$, we calculate their perplexity scores using equation \ref{ppl}, and select the response with the highest perplexity score as the challenging negative response for this conversation history. In practice, if all the scores are too small, we randomly sample an utterance as the negative response.

Through the whole generation process, we can generate one challenging negative response for each conversation history, we add these negative responses into the original training dataset, getting a new training set 1.5x larger than the original one (positive: negative = 1: 2). At last, we train the response selection model with a binary classification objective on the new training set.

\section{Experimental Setup}

\subsection{Datasets}
We evaluate our model on two widely used datasets: \textit{Douban Corpus} \cite{wu2016sequential}, and \textit{E-Commerce Corpus} \cite{zhang2018modeling}. All datasets consists of multi-turn conversations, and their statistics are summarized in Table \ref{dataset_info}.

\paragraph{Douban Corpus} consists of Chinese multi-turn daily conversations crawled from Douban website~\footnote{https://www.douban.com} which is a popular social networking service.

\paragraph{E-Commerce Corpus} is another Chinese multi-turn conversation corpus, it consists of conversations between customers and customer service staff from Taobao~\footnote{https://www.taobao.com}. The conversation talks about several types of topics, such as commodity recommendation, consultation, and negotiation.

\subsection{Evaluation Metrics} 
Following prior works \cite{lowe2015ubuntu, wu2016sequential, zhou2018multi, whang2020response}, we utilize several retrieval metrics to evaluate our negative response generation approach. 
For all the two datasets, we utilize $R_n@k(k={1,2,5})$, that is 1 in $n$ recall at $k$, it gets 1 if a golden response is positioned in the $k$ selected responses and 0 otherwise, for these two datasets, $n$ is equal to 10. 
Besides, for the Douban dataset, we also employ the other three metrics MAP(mean average), MRR(mean reciprocal rank), and P@1(precision at one) to evaluate the model's performance, since there may be more than one positive response among the candidates.

\begin{table*}[]
\centering
\begin{tabular}{l| llllll| lll}
\hline 
 & \multicolumn{6}{c|}{Douban} & \multicolumn{3}{c}{Ecommerce} \\
 & MAP & MRR & P1 & R1 & R2 & R5 & R1 & R2 & R5 \\
\hline 
ELECTRA       & 0.602  & 0.642 & 0.465 & 0.287 & 0.483 & 0.839 & 0.609      & 0.804 & 0.965  \\
ELECTRA+      & 0.612  & 0.655 & 0.480 & 0.301 & 0.499 & 0.836 & 0.673      & 0.835 & 0.974  \\
BERT & 0.591  & 0.633 & 0.454 & 0.280 & 0.470 & 0.828 & 0.610      & 0.814 & 0.973  \\
BERT+         & 0.609  & 0.645 & 0.463 & 0.290 & 0.505 & 0.838 & 0.725      & 0.890 & 0.984  \\
SA-BERT       & 0.619  & 0.659 & 0.496 & 0.313 & 0.481 & 0.847 & 0.704      & 0.879 & 0.985  \\
\hline
BIN-BERT-gen  & 0.565  & 0.607 & 0.424 & 0.264 & 0.431 & 0.807 & -          & -     & -      \\
BIN-BERT-ret  & 0.592  & 0.632 & 0.441 & 0.273 & 0.480 & 0.833 & -          & -     & -      \\
\hline
GGS-ELECTRA   &  0.611       &  0.651     &  0.483     &   0.305    &  0.484     &    0.836   &       0.651     &    0.829   &   0.975     \\
GGS-ELECTRA+  &   0.617     &    0.658   &    0.490   &   0.308    &    0.490   &    \textbf{0.848}   &     0.705       &    0.860   &   0.976    \\
GGS-BERT      &  0.596      &   0.637    &  0.459     &   0.282    & 0.475      &  0.833     &    0.661     &     0.847  &   0.977     \\
GGS-BERT+     &   \textbf{0.631}     &  \textbf{0.669}     &      \textbf{0.504} & \textbf{0.321}      &    \textbf{0.519}   &   0.834     &      \textbf{0.754}      &   \textbf{0.900}    &  \textbf{0.989}      \\
\hline
\end{tabular}
\caption{Evaluation results on Douban and E-Commerce datasets. Models named GGS-XXX are trained with challenging negative instances which leading to remarkable performance improvement consistently.}
\label{main_exp_result}
\end{table*}

\subsection{Baselines}
\paragraph{BERT-based Models.}
Pre-trained language models, such as BERT~\cite{devlin2018bert}, ELECTRA~\cite{clark2020electra} and their variants like SA-BERT~\cite{gu2020speaker} have been applied to the response selection task. 
In these models, conversation history and each candidate response are concatenated to a long sequence, and [CLS] token's hidden state on the last layer is used as the sequence representation, which is then passed into a binary classifier to predict whether the candidate response is correct. Following \newcite{whang2020response} we also post-training BERT and ELECTRA on the task-specific datasets using masked language model object before fine-tuning, these models are denoted as BERT+ and ELECTRA+.

\paragraph{Data Augmentation Approaches.} 
In the original training set of three datasets, the ratio of positive and negative instances is 1:1, the negative instances are randomly sampled, so there is a lot of room for data augmentation.
\newcite{lin2020world} leveraged retrieval model and seq2Seq generation model to construct greyscale data, and proposed multi-level ranking objectives to let the model learn a progressive relationship, such as \textit{"golden $>$ greyscale $>$ random"}. We utilize their retrieved and generated responses as the augmented negative instances for training BERT, the experiments are denoted as BIN-BERT-gen and BIN-BERT-ret.

\subsection{Training Details}
For negative response generation, we first fine-tune the pre-trained Cdial-GPT on two datasets individually with the learning rate equal to 5e-5, then we utilize the top-p sampling decoding method to generate responses according to garbled conversation history.
For the response selection model, we use BERT and ELECTRA implemented from huggingface Transformers\footnote{https://github.com/huggingface/transformers} based on PyTorch framework~\cite{paszke2019pytorch}. 
We initialize model parameters from checkpoints bert-base-chinese-wwm(the whole-word masking strategy) and electra-base-chinese
We also utilize the checkpoints released by \newcite{whang2020response}, which is post trained on these two datasets separately by masked language model objective. 
For fine-tuning stage, we train the models with a batch size of 32, a learning rate of 3e-5 using the Adam optimizer with linear learning rate decay, the max sequence length is 512. We run all the experiments on four Tesla V100 GPUs.

\section{Results and Analysis}

\subsection{Experiment Results}

The main experimental results are shown in Table \ref{main_exp_result}. The results of our approach is named as GGS-XXX, it means training the model using our challenging negative responses together with the original training set. 
To prove the effectiveness of our proposed method, we compare our method with strong pre-trained models such as BERT and their variants like SA-BERT. Besides, we also compare our method with other data augmentation methods such as BIN-BERT-ret and BIN-BERT-gen.

On all datasets, the models trained with our challenging negative response significantly outperform the models trained on the original dataset. 
The performance improvements are consistent across different datasets and different models, indicating the universality of our proposed method.
In addition, when applied to a high-accurate model, such as GGS-BERT+ on E-Commerce dataset, our approach can still achieve remarkable performance improvement over all evaluation metrics. 

Besides, GGS-BERT outperforms other data augmentation methods based on BERT model, which also indicates the effectiveness of our proposed generation method and the value of challenging negative instances.
It is worth noting that BIN-BERT-gen makes the model's performance drop. Through the case study, we found that many of the responses generated by the seq2seq model are false negative, which also reflects that it is not reliable to simply generate negative samples with the generation model, and a filter needs to be added.

\subsection{Ablation Study}
We conduct ablation studies for investigating the effect of different modules. We choose BERT and BERT+ as the base models and train them on E-Commerce dataset with five additional settings.

\begin{table}[]
\centering
\small
\begin{tabular}{l l l l}
\hline
 & Recall@1 & Recall@2 & Recall@5 \\
\hline
GGS-BERT & 0.661 & 0.847 & 0.977 \\
GGS-BERT+ & 0.754 & 0.900 & 0.989 \\
\begin{tabular}[c]{@{}l@{}}GGS-BERT\\  \textit{w.o. garble} \end{tabular} & 0.431↓ & 0.690↓ & 0.954↓ \\
\begin{tabular}[c]{@{}l@{}}GGS-BERT+\\   \textit{w.o. garbled} \end{tabular} & 0.568↓ & 0.795↓ & 0.980↓ \\
\hline 
\end{tabular}
\caption{The results of ablation experiment removing garbled history construction module on E-Commerce dataset. The performance significantly drops, even worse than the baselines.}
\label{ablation_garble}
\end{table}

\paragraph{Effect of Garbled History.}
In this setting, we feed the original conversation history instead of the garbled history to Cdial-GPT to generate responses and use them as negative training instances. 
The experiment results listed in Table~\ref{ablation_garble} shows that removing the history garbling mechanism makes the models' performance drop significantly, even much worse than baseline.
This indicates that the garbled history construction module makes an irreplaceable contribution to the generation of challenging negative instances. It is because Cdial-GPT is pre-trained on large-scale open-domain dialogue corpus, it is likely to generate a coherent and natural response, most of which could be fake negative.

\begin{table}[]
\centering
\small
\begin{tabular}{llll}
\hline
 & Recall@1 & Recall@2 & Recall@5 \\
\hline
GGS-BERT & 0.661 & 0.847 & 0.977 \\
GGS-BERT+ & 0.754 & 0.900 & 0.989 \\
\begin{tabular}[c]{@{}l@{}}GGS-BERT\\   \textit{w.o. filter}\end{tabular} & 0.622↓ & 0.822↓ & 0.970↓ \\
\begin{tabular}[c]{@{}l@{}}GGS-BERT+\\   \textit{w.o. filter}\end{tabular} & 0.729↓ & 0.887↓ & 0.985↓ \\
\hline
\end{tabular}
\caption{The results of ablation experiment removing response selection module on E-Commerce dataset. The performance drops across all metrics.}
\label{ablation_filter}
\end{table}

\paragraph{Effect of Response Selection.}
Although the garbled conversation history can prevent Cdial-GPT to generate coherent and natural responses to a large extent, considering the diversity characteristic of the open-domain dialogue data, we still propose a selector to select responses that are more likely to be negative by ranking their perplexity score based on the original conversation history. 
We study the effect of the selector by replacing it with a random response selector. 

Experiment results are listed in Table~\ref{ablation_filter}. Removing the response selection mechanism brings some performance drop, indicating the effectiveness of the proposed selector. Meanwhile, the extent of the performance drop is not as significant as removing garbled history, indicating garbled history is more important to avoid fake negative responses.

\begin{table}[]
\centering
\small
\begin{tabular}{llll}
\hline 
 & Recall@1 & Recall@2 & Recall@5 \\
 \hline
GGS-BERT & 0.661 & 0.847 & 0.977 \\
GGS-BERT+ & 0.754 & 0.900 & 0.989 \\
\begin{tabular}[c]{@{}l@{}}GGS-BERT\\  \textit{w.o. gen1}\end{tabular} & 0.616↓ & 0.834↓ & 0.979 \\
\begin{tabular}[c]{@{}l@{}}GGS-BERT\\  \textit{w.o. gen2}\end{tabular} & 0.620↓ & 0.812↓ & 0.967↓ \\
\begin{tabular}[c]{@{}l@{}}GGS-BERT+\\  \textit{w.o. gen1}\end{tabular} & 0.726↓ & 0.886↓ & 0.989 \\
\begin{tabular}[c]{@{}l@{}}GGS-BERT+\\  \textit{w.o. gen2}\end{tabular} & 0.691↓ & 0.879↓ & 0.987↓ \\
\hline 
\end{tabular}
\caption{The results of the ablation experiment investigating the contribution of two different generation methods. Removing any of them leads to a performance drop, while the drop of removing $gen2$ is greater.}
\label{ablation_gen}
\end{table}
\begin{table*}[]
\centering
\small
\begin{tabular}{p{7cm}  p{7cm}}
\hline 
\textbf{Conversation history} & \\
\begin{CJK}{UTF8}{gbsn} X: 你们应该质量上也是这样吧？\end{CJK} & X: Should the quality be ok? \\
\begin{CJK}{UTF8}{gbsn}Y: 不用担心质量方面，小店有保证的。\end{CJK} & Y: Don’t worry about the quality, it is guaranteed by our store. \\
\begin{CJK}{UTF8}{gbsn}X: 好甜吗？\end{CJK} & X: Is it sweet? \\
\begin{CJK}{UTF8}{gbsn}Y: 口感是偏甜的，湿度适中，可以当零食吃的。\end{CJK} & Y: It tastes sweet, has moderate moisture content, and can be eaten as snacks. \\
\begin{CJK}{UTF8}{gbsn}X: 我去拍。\end{CJK} & X: I’m going to place an order. \\
\begin{CJK}{UTF8}{gbsn}Y: 亲请核对一下收货地址哦。\end{CJK} & Y: Please check the shipping address. \\
\begin{CJK}{UTF8}{gbsn}X: 对，谢谢。 \end{CJK}& X: It’s correct, thanks. \\
\hline 
\textbf{Responses} & \\
Golden: \begin{CJK}{UTF8}{gbsn}小店尽快给您发出哦。 \end{CJK}& Golden: We will send it to you as soon as possible. \\
Random: \begin{CJK}{UTF8}{gbsn}不能哦。 \end{CJK}& Random: No, you can’t. \\
Challenging 1: \begin{CJK}{UTF8}{gbsn}收货地址没有改成功。 \end{CJK}& Challenging 1:  Shipping address wasn’t changed successfully. \\
Challenging 2: \begin{CJK}{UTF8}{gbsn}补发的单号会留言给亲收货地址。 \end{CJK}&Challenging 2:  The replacement order number will be sent to the shipping address. \\
Challenging 3: \begin{CJK}{UTF8}{gbsn}口感偏甜，回味带酸，回味很带核桃味哦。 \end{CJK}& Challenging 3: The taste is sweet with sour and walnut aftertaste. \\
\hline 
\end{tabular}
\caption{A case from E-Commerce corpus. The original dialogue is in Chinese(the left), we also provide their English version(the right). Golden and random responses are directly copied from the original dataset. The three challenging responses are generated by our proposed method.}
\label{case_study}
\end{table*}

\paragraph{Effect of Different Generation Methods.}
As shown in Figure~\ref{main_pic}, there are two negative response generation methods. 
1) Garble conversation history using flow distortion strategy first and then utilize Cdial-GPT directly generate the response based on the garbled history, denoted as \textit{gen1}. 
2) Use context destruction strategy to garble the conversation history, and generate responses that must contain the keyword of the original conversation history, denoted as \textit{gen2}. 

We study each method's effect. Experiment results are listed in Table~\ref{ablation_gen}, removing any of these two methods causes the model's performance to drop, indicating the effectiveness of them. 
Besides, removing \textit{gen2} leads to more performance drops indicating that \textit{gen2} contributes more than \textit{gen1}, It is because responses generated by \textit{gen2} must contain a keyword which is also mentioned in the original conversation history, making the responses more confusing to the model.

\paragraph{Effect of Challenging Instances.}
We replace all challenging responses with random sampled negative responses, the experiment is denoted as \textit{random DA} (DA is short for data augmentation). 
The experiment results are shown in Table~\ref{ablation_challenging}, 
replacing challenging instances with randomly sampled instances leads to a performance drop, indicating that the challenging instances are worth learning in the response selection task. 
In addition, the performance of \textit{BERT+ with random DA} is worse than \textit{BERT+}, it indicates that just adding random sampled negative instances to the training set does not necessarily improve the performance, it may be because E-Commerce corpus is domain-specific, it's mainly about commodity, randomly sampled instances are more likely to be fake negative.

\subsection{Case Study}
Table~\ref{case_study} shows a case from the E-Commerce corpus. 
The randomly sampled response is totally irrelevant to the conversation history, so it is easy to be distinguished by the model. 
Though the challenging responses generated by our method are not the proper responses because they have little coherence with the dialogue context. But they are quite confusing and challenging because they have some overlapped words or phrases with the conversation history, such as \textit{shipping address} and \textit{taste}.
The model trained on the original training dataset tends to incorrectly classify the challenging instances to be positive. In contrast, after learning from the challenging instances, models can identify the improper pattern in the challenging negative responses.

\begin{table}[]
\centering
\small
\begin{tabular}{l l ll}
\hline 
 & Recall@1 & Recall@2 & Recall@5 \\
\hline
GGS-BERT & 0.661 & 0.847 & 0.977 \\
GGS-BERT+ & 0.754 & 0.900 & 0.989 \\
\begin{tabular}[c]{@{}l@{}}BERT\\  \textit{random DA}\end{tabular} & 0.614↓ & 0.819↓ & 0.972↓ \\
\begin{tabular}[c]{@{}l@{}}BERT+\\  \textit{random DA}\end{tabular} & 0.689↓ & 0.881↓ & 0.982↓ \\
\hline
\end{tabular}
\caption{Ablation experiment investigating the effect of challenging instances. Replacing challenging instances with random instances leads to a performance drop.}
\label{ablation_challenging}
\end{table}

\section{Conclusion}

We propose a challenging instances generation method leveraging the large scale pre-trained language models for enhancing dialogue response selection models. 
Firstly, we garble the conversation history with two strategies to address the fake negative issue, then a fine-tuned DialoGPT is leveraged to generate response according to the garbled history using two decoding methods, at last, we utilize a response selector to choose the most negative responses.
Compared to randomly sampled negatives, the challenging negative instances are more confusing and difficult to be distinguished which can help the model to learn more general and robust patterns about the natural dialogue response.
Experiment results on two benchmark datasets and four models demonstrate that the challenging instances are valuable for training response selection models.
Since our method is independent of model structure, we will apply it to more competitive response selection models in future work.

\bibliography{anthology,custom}

\begin{thebibliography}{26}
\expandafter\ifx\csname natexlab\endcsname\relax\def\natexlab#1{#1}\fi

\bibitem[{Clark et~al.(2020)Clark, Luong, Le, and Manning}]{clark2020electra}
Kevin Clark, Minh{-}Thang Luong, Quoc~V. Le, and Christopher~D. Manning. 2020.
\newblock \href {https://openreview.net/forum?id=r1xMH1BtvB} {{ELECTRA:}
  pre-training text encoders as discriminators rather than generators}.
\newblock In \emph{8th International Conference on Learning Representations,
  {ICLR} 2020, Addis Ababa, Ethiopia, April 26-30, 2020}. OpenReview.net.

\bibitem[{Devlin et~al.(2019)Devlin, Chang, Lee, and
  Toutanova}]{devlin2018bert}
Jacob Devlin, Ming-Wei Chang, Kenton Lee, and Kristina Toutanova. 2019.
\newblock Bert: Pre-training of deep bidirectional transformers for language
  understanding.
\newblock In \emph{Proceedings of the 2019 Conference of the North American
  Chapter of the Association for Computational Linguistics: Human Language
  Technologies, Volume 1 (Long and Short Papers)}, pages 4171--4186.

\bibitem[{Gu et~al.(2020)Gu, Li, Liu, Ling, Su, Wei, and Zhu}]{gu2020speaker}
Jia-Chen Gu, Tianda Li, Quan Liu, Zhen-Hua Ling, Zhiming Su, Si~Wei, and
  Xiaodan Zhu. 2020.
\newblock Speaker-aware bert for multi-turn response selection in
  retrieval-based chatbots.
\newblock In \emph{Proceedings of the 29th ACM International Conference on
  Information \& Knowledge Management}, pages 2041--2044.

\bibitem[{Humeau et~al.(2020)Humeau, Shuster, Lachaux, and
  Weston}]{humeau2019poly}
Samuel Humeau, Kurt Shuster, Marie{-}Anne Lachaux, and Jason Weston. 2020.
\newblock \href {https://openreview.net/forum?id=SkxgnnNFvH} {Poly-encoders:
  Architectures and pre-training strategies for fast and accurate
  multi-sentence scoring}.
\newblock In \emph{8th International Conference on Learning Representations,
  {ICLR} 2020, Addis Ababa, Ethiopia, April 26-30, 2020}. OpenReview.net.

\bibitem[{Kim et~al.(2020)Kim, Eric, Gopalakrishnan, Hedayatnia, Liu, and
  Hakkani-Tur}]{kim2020beyond}
Seokhwan Kim, Mihail Eric, Karthik Gopalakrishnan, Behnam Hedayatnia, Yang Liu,
  and Dilek Hakkani-Tur. 2020.
\newblock Beyond domain apis: Task-oriented conversational modeling with
  unstructured knowledge access.
\newblock In \emph{Proceedings of the 21th Annual Meeting of the Special
  Interest Group on Discourse and Dialogue}, pages 278--289.

\bibitem[{Kollar et~al.(2018)Kollar, Berry, Stuart, Owczarzak, Chung, Mathias,
  Kayser, Snow, and Matsoukas}]{kollar2018alexa}
Thomas Kollar, Danielle Berry, Lauren Stuart, Karolina Owczarzak, Tagyoung
  Chung, Lambert Mathias, Michael Kayser, Bradford Snow, and Spyros Matsoukas.
  2018.
\newblock The alexa meaning representation language.
\newblock In \emph{Proceedings of the 2018 Conference of the North American
  Chapter of the Association for Computational Linguistics: Human Language
  Technologies, Volume 3 (Industry Papers)}, pages 177--184.

\bibitem[{Lin et~al.(2020)Lin, Cai, Wang, Liu, Zheng, and Shi}]{lin2020world}
Zibo Lin, Deng Cai, Yan Wang, Xiaojiang Liu, Haitao Zheng, and Shuming Shi.
  2020.
\newblock The world is not binary: Learning to rank with grayscale data for
  dialogue response selection.
\newblock In \emph{Proceedings of the 2020 Conference on Empirical Methods in
  Natural Language Processing (EMNLP)}, pages 9220--9229.

\bibitem[{Liu et~al.(2019)Liu, Ott, Goyal, Du, Joshi, Chen, Levy, Lewis,
  Zettlemoyer, and Stoyanov}]{liu2019roberta}
Yinhan Liu, Myle Ott, Naman Goyal, Jingfei Du, Mandar Joshi, Danqi Chen, Omer
  Levy, Mike Lewis, Luke Zettlemoyer, and Veselin Stoyanov. 2019.
\newblock \href {http://arxiv.org/abs/1907.11692} {Roberta: {A} robustly
  optimized {BERT} pretraining approach}.
\newblock \emph{CoRR}, abs/1907.11692.

\bibitem[{Lowe et~al.(2015)Lowe, Pow, Serban, and Pineau}]{lowe2015ubuntu}
Ryan Lowe, Nissan Pow, Iulian~Vlad Serban, and Joelle Pineau. 2015.
\newblock The ubuntu dialogue corpus: A large dataset for research in
  unstructured multi-turn dialogue systems.
\newblock In \emph{Proceedings of the 16th Annual Meeting of the Special
  Interest Group on Discourse and Dialogue}, pages 285--294.

\bibitem[{Lu et~al.(2020)Lu, Ren, Ren, Liu, and Xu}]{lu2020improving}
Junyu Lu, Xiancong Ren, Yazhou Ren, Ao~Liu, and Zenglin Xu. 2020.
\newblock Improving contextual language models for response retrieval in
  multi-turn conversation.
\newblock In \emph{Proceedings of the 43rd International ACM SIGIR Conference
  on Research and Development in Information Retrieval}, pages 1805--1808.

\bibitem[{Lu et~al.(2019)Lu, Zhang, Xie, Ling, Zhou, and
  Xu}]{lu2019constructing}
Junyu Lu, Chenbin Zhang, Zeying Xie, Guang Ling, Tom~Chao Zhou, and Zenglin Xu.
  2019.
\newblock Constructing interpretive spatio-temporal features for multi-turn
  responses selection.
\newblock In \emph{Proceedings of the 57th Annual Meeting of the Association
  for Computational Linguistics}, pages 44--50.

\bibitem[{Paszke et~al.(2019)Paszke, Gross, Massa, Lerer, Bradbury, Chanan,
  Killeen, Lin, Gimelshein, Antiga et~al.}]{paszke2019pytorch}
Adam Paszke, Sam Gross, Francisco Massa, Adam Lerer, James Bradbury, Gregory
  Chanan, Trevor Killeen, Zeming Lin, Natalia Gimelshein, Luca Antiga, et~al.
  2019.
\newblock Pytorch: An imperative style, high-performance deep learning library.
\newblock \emph{Advances in Neural Information Processing Systems},
  32:8026--8037.

\bibitem[{Radford et~al.(2019)Radford, Wu, Child, Luan, Amodei, and
  Sutskever}]{radford2019language}
Alec Radford, Jeffrey Wu, Rewon Child, David Luan, Dario Amodei, and Ilya
  Sutskever. 2019.
\newblock Language models are unsupervised multitask learners.
\newblock \emph{OpenAI blog}, 1(8):9.

\bibitem[{Raffel et~al.(2020)Raffel, Shazeer, Roberts, Lee, Narang, Matena,
  Zhou, Li, and Liu}]{raffel2019exploring}
Colin Raffel, Noam Shazeer, Adam Roberts, Katherine Lee, Sharan Narang, Michael
  Matena, Yanqi Zhou, Wei Li, and Peter~J Liu. 2020.
\newblock Exploring the limits of transfer learning with a unified text-to-text
  transformer.
\newblock \emph{Journal of Machine Learning Research}, 21:1--67.

\bibitem[{Shum et~al.(2018)Shum, He, and Li}]{shum2018eliza}
Heung-Yeung Shum, Xiao-dong He, and Di~Li. 2018.
\newblock From eliza to xiaoice: challenges and opportunities with social
  chatbots.
\newblock \emph{Frontiers of Information Technology \& Electronic Engineering},
  19(1):10--26.

\bibitem[{Tao et~al.(2019)Tao, Wu, Xu, Hu, Zhao, and Yan}]{tao2019multi}
Chongyang Tao, Wei Wu, Can Xu, Wenpeng Hu, Dongyan Zhao, and Rui Yan. 2019.
\newblock Multi-representation fusion network for multi-turn response selection
  in retrieval-based chatbots.
\newblock In \emph{Proceedings of the Twelfth ACM International Conference on
  Web Search and Data Mining}, pages 267--275.

\bibitem[{Vaswani et~al.(2017)Vaswani, Shazeer, Parmar, Uszkoreit, Jones,
  Gomez, Kaiser, and Polosukhin}]{vaswani2017attention}
Ashish Vaswani, Noam Shazeer, Niki Parmar, Jakob Uszkoreit, Llion Jones,
  Aidan~N Gomez, {\L}ukasz Kaiser, and Illia Polosukhin. 2017.
\newblock Attention is all you need.
\newblock In \emph{Advances in neural information processing systems}, pages
  5998--6008.

\bibitem[{Wang et~al.(2020)Wang, Ke, Zheng, Huang, Jiang, Zhu, and
  Huang}]{wang2020chinese}
Yida Wang, Pei Ke, Yinhe Zheng, Kaili Huang, Yong Jiang, Xiaoyan Zhu, and
  Minlie Huang. 2020.
\newblock \href {https://arxiv.org/abs/2008.03946} {A large-scale chinese
  short-text conversation dataset}.
\newblock In \emph{NLPCC}.

\bibitem[{Whang et~al.(2020)Whang, Lee, Lee, Yang, Oh, and
  Lim}]{whang2019effective}
Taesun Whang, Dongyub Lee, Chanhee Lee, Kisu Yang, Dongsuk Oh, and Heuiseok
  Lim. 2020.
\newblock An effective domain adaptive post-training method for bert in
  response selection.
\newblock \emph{Proc. Interspeech 2020}, pages 1585--1589.

\bibitem[{Whang et~al.(2021)Whang, Lee, Oh, Lee, Han, Lee, and
  Lee}]{whang2020response}
Taesun Whang, Dongyub Lee, Dongsuk Oh, Chanhee Lee, Kijong Han, Dong-hun Lee,
  and Saebyeok Lee. 2021.
\newblock Do response selection models really know what's next? utterance
  manipulation strategies for multi-turn response selection.
\newblock In \emph{Proceedings of the AAAI Conference on Artificial
  Intelligence}.

\bibitem[{Wu et~al.(2017)Wu, Wu, Xing, Zhou, and Li}]{wu2016sequential}
Yu~Wu, Wei Wu, Chen Xing, Ming Zhou, and Zhoujun Li. 2017.
\newblock Sequential matching network: A new architecture for multi-turn
  response selection in retrieval-based chatbots.
\newblock In \emph{Proceedings of the 55th Annual Meeting of the Association
  for Computational Linguistics (Volume 1: Long Papers)}, pages 496--505.

\bibitem[{Yuan et~al.(2019)Yuan, Zhou, Li, Lv, Zhu, Han, and
  Hu}]{yuan2019multi}
Chunyuan Yuan, Wei Zhou, Mingming Li, Shangwen Lv, Fuqing Zhu, Jizhong Han, and
  Songlin Hu. 2019.
\newblock Multi-hop selector network for multi-turn response selection in
  retrieval-based chatbots.
\newblock In \emph{Proceedings of the 2019 Conference on Empirical Methods in
  Natural Language Processing and the 9th International Joint Conference on
  Natural Language Processing (EMNLP-IJCNLP)}, pages 111--120.

\bibitem[{Zhang et~al.(2020)Zhang, Sun, Galley, Chen, Brockett, Gao, Gao, Liu,
  and Dolan}]{zhang2019dialogpt}
Yizhe Zhang, Siqi Sun, Michel Galley, Yen-Chun Chen, Chris Brockett, Xiang Gao,
  Jianfeng Gao, Jingjing Liu, and William~B Dolan. 2020.
\newblock Dialogpt: Large-scale generative pre-training for conversational
  response generation.
\newblock In \emph{Proceedings of the 58th Annual Meeting of the Association
  for Computational Linguistics: System Demonstrations}, pages 270--278.

\bibitem[{Zhang et~al.(2018)Zhang, Li, Zhu, Zhao, and Liu}]{zhang2018modeling}
Zhuosheng Zhang, Jiangtong Li, Pengfei Zhu, Hai Zhao, and Gongshen Liu. 2018.
\newblock Modeling multi-turn conversation with deep utterance aggregation.
\newblock In \emph{Proceedings of the 27th International Conference on
  Computational Linguistics}, pages 3740--3752.

\bibitem[{Zhou et~al.(2016)Zhou, Dong, Wu, Zhao, Yu, Tian, Liu, and
  Yan}]{zhou2016multi}
Xiangyang Zhou, Daxiang Dong, Hua Wu, Shiqi Zhao, Dianhai Yu, Hao Tian, Xuan
  Liu, and Rui Yan. 2016.
\newblock Multi-view response selection for human-computer conversation.
\newblock In \emph{Proceedings of the 2016 Conference on Empirical Methods in
  Natural Language Processing}, pages 372--381.

\bibitem[{Zhou et~al.(2018)Zhou, Li, Dong, Liu, Chen, Zhao, Yu, and
  Wu}]{zhou2018multi}
Xiangyang Zhou, Lu~Li, Daxiang Dong, Yi~Liu, Ying Chen, Wayne~Xin Zhao, Dianhai
  Yu, and Hua Wu. 2018.
\newblock Multi-turn response selection for chatbots with deep attention
  matching network.
\newblock In \emph{Proceedings of the 56th Annual Meeting of the Association
  for Computational Linguistics (Volume 1: Long Papers)}, pages 1118--1127.

\end{thebibliography}
\bibliographystyle{acl_natbib}

\end{document}